\documentclass[11pt,a4paper]{article}
\usepackage[hyperref]{acl2021}
\usepackage{times}
\usepackage{latexsym}
\usepackage{float}

\usepackage{graphicx}
\usepackage{capt-of}
\usepackage{amsmath}
\usepackage{subcaption}
\usepackage[section]{placeins}
\usepackage{todonotes}
\usepackage{fixfoot}
\usepackage{stfloats}
\graphicspath{ {./images/} }

\usepackage{microtype}

\aclfinalcopy 
\def\aclpaperid{46} 

\title{Zero-shot Sequence Labeling for Transformer-based Sentence Classifiers}

\author{Kamil Bujel \\
  Department of Computing \\
  Imperial College London \\
  United Kingdom \\
  \texttt{\footnotesize kdb19@imperial.ac.uk} \\\And
  Helen Yannakoudakis \\
  Department of Informatics \\
  King's College London \\
  United Kingdom \\
  \texttt{\footnotesize helen.yannakoudakis@kcl.ac.uk} \\\And
  Marek Rei \\
  Department of Computing \\
  Imperial College London \\
  United Kingdom \\
  \texttt{\footnotesize marek.rei@imperial.ac.uk}
   \\}

\date{}

\begin{document}
\maketitle
\begin{abstract}
We investigate how sentence-level transformers can be modified into effective sequence labelers at the token level without any direct supervision. Existing approaches to zero-shot sequence labeling do not perform well when applied on transformer-based architectures. 
As transformers contain multiple layers of multi-head self-attention, information in the sentence gets distributed between many tokens, negatively affecting zero-shot token-level performance.  
We find that a soft attention module which explicitly encourages sharpness of attention weights can significantly outperform existing methods. 



 
\end{abstract}

\section{Introduction}

Sequence labeling and sentence classification can represent facets of the same task at different granularities; for example, detecting grammar errors and predicting the grammaticality of sentences. 
Transformer-based architectures such as BERT \citep{devlin-etal-2019-bert} and RoBERTa \citep{liu2019roberta} have been shown to achieve state-of-the-art results on both sequence labeling \citep{bell2019context} and sentence classification \citep{sun2019fine} problems. However, such tasks are typically treated in isolation rather than within a unified approach. 

In this paper, we investigate methods for inferring token-level predictions from transformer models trained only on sentence-level annotations. 
The ability to classify individual tokens without direct supervision opens possibilities for training sequence labeling models on tasks and datasets where only sentence-level or document-level annotation is available. 
In addition, attention-based architectures allow us to directly investigate what the model is learning and to quantitatively measure whether its \textit{rationales} (supporting evidence) for particular input sentences match human expectations. 
While evaluating the \textit{faithfulness} \citep{herman2017promise} of a model's rationale is still an open research question and up for debate \citep{jain-wallace-2019-attention, wiegreffe-pinter-2019-attention, deyoung-etal-2020-eraser, jacovi-goldberg-2020-towards, atanasova2020diagnostic}, the methods explored here allow for measuring the \textit{plausibility} (agreeability to human annotators; \citet{deyoung-etal-2020-eraser}) of transformer-based models using existing sequence labeling datasets. 


%

We evaluate and compare different methods for adapting pre-trained transformer models into zero-shot sequence labelers, trained using only gold sentence-level signal. 
Our experiments show that applying existing approaches \cite{rei-sogaard-2018-zero} to transformer architectures is not straightforward -- transformers already contain several layers of multi-head attention, distributing sentence-level information across many tokens, whereas the existing methods rely on all the information going through one central attention module. 
Approaches such as LIME \citep{ribeiro2016should} for scoring word importance 
also struggle to infer correct token-level annotations in a zero-shot manner (e.g., it achieves only $2\%$ F-score on one of our datasets). 
We find that a modified attention function is needed to allow transformers to better focus on individual important tokens and achieve a new state-of-the-art on zero-shot sequence labeling.






The contributions of this paper are fourfold: 
\begin{itemize}
\item We present the first experiments utilizing (pre-trained) sentence-level transformers as zero-shot sequence labelers; 

\item We perform a systematic comparison of alternative methods for zero-shot sequence labeling on different datasets; 

\item We propose a novel modification of the attention function that significantly improves zero-shot sequence-labeling performance of transformers over the previous state of the art, while achieving on-par or better results on sentence classification; 

\item We make our source code and models publicly available to facilitate further research in the field.\footnote{\url{https://github.com/bujol12/bert-seq-interpretability}}
\end{itemize}

\section{Methods}
We evaluate four different methods for turning sentence-level transformer models into zero-shot sequence labelers. 

\subsection{LIME}
LIME \citep{ribeiro2016should} generates local word-level importance scores through a meta-model that is trained on perturbed data generated by randomly masking out words in the input sentence. 
It was originally investigated in the context of Support Vector Machine \citep{hearst1998support} text classifiers with unigram features.

We apply LIME to a RoBERTa model supervised as a sentence classifier and investigate whether its scores can be used for sequence labeling.  We use RoBERTa's MASK token to mask out individual words and allow LIME to generate $5000$ masked samples per sentence. The resulting explanation weights are then used as classification scores for each word, with the decision threshold fine-tuned based on the development set performance.

\citet{thorne-etal-2019-generating} found LIME to outperform attention-based approaches on the task of explaining NLI models. LIME was used to probe a LSTM-based sentence-pair classifier \citep{lan-xu-2018-neural} by removing tokens from the premise and hypothesis sentences separately. The generated scores were used to perform binary classification of tokens, with the threshold based on $F_1$ performance on the development set. The token-level predictions were evaluated against human explanations of the entailment relation using the e-SNLI dataset \citep{camburu2018snli}. LIME was found to outperform other methods, however, it was also $1000\times$ slower than attention-based methods at generating these explanations. 

\subsection{Attention heads}
The attention heads in a trained transformer model are designed to identify and combine useful information for a particular task. 
\citet{clark-etal-2019-bert} found that specific heads can specialize on different linguistic properties such as syntax and coreference.
However, transformer models contain many layers with multiple attention heads, distributing the text representation and making it more difficult to identify token importance for the overall task.

Given a particular head, we can obtain an importance score for each token by averaging the attention scores from all the tokens that attend to it.
In order to investigate the best possible setting, we report results for the attention head that achieves the highest token-level Mean Average Precision score on the development set.


\subsection{Soft attention}

\citet{rei-sogaard-2018-zero} described a method for predicting token-level labels based on a bidirectional LSTM \citep{hochreiter1997long} architecture supervised at the sentence-level only. 
A dedicated attention module was integrated for building sentence representations, with its attention weights also acting as token-level importance scores. The architecture was found to outperform a gradient-based approach on the tasks of zero-shot sequence labeling for error detection, uncertainty detection, and sentiment analysis.

In order to obtain a single raw attention value $\widetilde{e}_i$ for each token, biLSTM output vectors were passed through a feedforward layer:
\begin{equation}
    e_i = tanh(W_e h_i + b_e) \;\;\;\; \widetilde{e}_i = W_{\widetilde{e}} e_i + b_{\widetilde{e}}
\end{equation}

\noindent where $e_i$ is the attention vector for token $t_i$; $h_i$ is the biLSTM output for $t_i$; and $\widetilde{e}_i$ is the single raw attention value. $W_e$, $b_e$, $W_{\widetilde{e}}$, $b_{\widetilde{e}}$ are trainable parameters. 

Instead of softmax or sparsemax \citep{martins2016softmax}, which would restrict the distribution of the scores, a soft attention based on sigmoid activation was used to obtain importance scores:

\begin{equation}
    \widetilde{a}_i = \sigma(\widetilde{e}_i) \;\;\;\; \;\;\;\;
    a_i = \frac{\widetilde{a}_i}{\sum_{k=1}^{N}\widetilde{a}_k}
\end{equation}

\noindent where $N$ is the number of tokens and $\sigma$ is the logistic function. $\widetilde{a}_i$ shows the importance of a particular token and is in the range $0 \leq \widetilde{a}_i \leq 1$, independent of any other scores in the sentence; therefore, it can be directly used for sequence labeling with a natural threshold of $0.5$.
$a_i$ contains the same information but is normalized to sum up to $1$ over the whole sentence, making it suitable for attention weights when building the sentence representation. As $a_i$ and $\widetilde{a}_i$ are directly tied, training the former through the sentence classification objective will also train the latter for the sequence labeling task.

The attention values were then used to obtain the sentence representation $c$ by acting as weights for the biLSTM token outputs:

\begin{equation}
    c = \sum_{i=0}^{N} a_i h_i
\end{equation}

Finally, the sentence representation $c$ was passed through the final feedforward layer, followed by a sigmoid to obtain the predicted score $y$ for the sentence:
\begin{equation}
    d = tanh(W_d c + b_d) \;\;\;\; \;\; y = \sigma(W_y d + b_y)
\end{equation}

\noindent where $d$ is the sentence vector, $c$ is the sentence representation, and $y$ is the sentence prediction score. $W_d$, $b_d$, $W_y$, $b_y$ are all trainable parameters. 

We adapt this approach to the transformer models by attaching a separate soft attention module on top of the token-level output representations. This effectively ignores the CLS token, which is commonly used for sentence classification, and instead builds a new sentence representation from the token representations, which replace the previously used biLSTM outputs:
\begin{equation}
    e_i = tanh(W_e T_i + b_e) \;\;\;\; c = \sum_{i=0}^{N} a_i T_i
\end{equation}

\noindent where $T_i$ is the contextualized embedding for token $t_i$. 
A diagram of the model architecture is included in Appendix \ref{fig:sa-diag}.

Commonly used tokenizers for transformer models split words into subwords, while sequence labeling datasets are annotated at the word level. 
We find that taking the maximum attention value over all the subwords as the word-level importance score produces good results on the development sets. For a word $w_i$ split into tokens $[t_j, ..., t_m]$, where $j,m \in [1, N]$, the resulting final word importance score $r_i$ is then given by:
\begin{equation}
    \label{combine_tokens}
    r_i = max(\{\widetilde{a}_j, \widetilde{a}_{j+1}, ..., \widetilde{a}_m\})
\end{equation}

During training, we optimize sentence-level binary cross-entropy as the main objective function:
\begin{equation}
    L_1 = \frac{\sum_j CrossEntropy(y^{(j)}, \widetilde{y}^{(j)})} {|y|}
\end{equation}
\noindent where $y^{(j)}$ and $\tilde{y}^{(j)}$ are the predicted sentence classification logits and the gold label for the $j^{th}$ sentence respectively. 
We also adopt the additional loss functions from \citet{rei-sogaard-2018-zero}, which encourage the attention weights to behave more like token-level classifiers:

\begin{equation}
    \label{l2}
    L_2 = \frac{\sum_j (min_j(\widetilde{a_i}) - 0)^2} {|y|}
\end{equation}
\begin{equation}
    \label{l3}
    L_3 = \frac{\sum_j (max_j(\widetilde{a_i}) - \widetilde{y}^{(j)})^2} {|y|}
\end{equation}

\noindent Eq. \ref{l2} optimizes the minimum unnormalized attention to be $0$ and therefore incentivizes the model to only focus on some, but not all words; Eq. \ref{l3} ensures that some attention weights are close to $1$ if the overall sentence is classified as positive. 
We then jointly optimize these three loss functions using a hyperparameter $\gamma$: $ L = L_1 + \gamma (L_2 + L_3)$.

\subsection{Weighted soft attention}
\label{wsa}
Our experiments show that, when combined with transformer-based models, the soft attention method tends to spread out the attention too widely. 
Instead of focusing on specific important words, the model broadly attends to the whole sentence.
Figures \ref{fig:sfig1} and \ref{fig:sfig3} in Appendix \ref{ap:eg} present examples demonstrating such behaviour. 
As transformers contain several layers of attention, with multiple heads in each layer, the information in the sentence gets distributed across all tokens before it reaches the soft attention module at the top.

To improve this behaviour and incentivize the model to direct information through a smaller and more focused set of tokens, we experiment with a weighted soft attention: 
\begin{equation}
    \label{attn_norm}
    a_i = \frac{\tilde{a}_i^\beta}{\sum_{k=1}^{N}\tilde{a}_k^\beta}
\end{equation}

\noindent where $\beta$ is a hyperparamete and where values $\beta > 1$ make the weight distribution sharper, allowing the model to focus on a smaller number of tokens. 
We experiment with values of $\beta \in \{1, 2, 3, 4\}$ on the development sets and find $\beta = 2$ to significantly improve token labeling performance without negatively affecting sentence classification results.




\section{Datasets}\label{datasets}

We investigate the performance of these methods as zero-shot sequence labelers using three different datasets. Gold token-level annotation in these datasets is used for evaluation; however, the models are trained using sentence-level labels only.

The \textbf{CoNLL 2010} shared task \citep{farkas-etal-2010-conll}\footnote{\url{https://rgai.sed.hu/node/118}} focuses on the detection of uncertainty cues in natural language text. The dataset contains $19,542$ examples with both sentence-level uncertainty labels and annotated keywords indicating uncertainty. 
We use the train/test data from the task and randomly choose $10\%$ of the training set for development.

We also evaluate on the task of grammatical error detection (GED) -- identifying which sentences are grammatically incorrect (i.e., contain at least one grammatical error). 
The First Certificate in English dataset \textbf{FCE} \cite{yannakoudakis-etal-2011-new}  consists of essays written by non-native learners of English, annotated for grammatical errors. We use the train/dev/test splits released by \citet{rei-yannakoudakis-2016-compositional} for sequence labeling, with a total of $33,673$ sentences. 

In addition, we evaluate on the Write \& Improve \cite{yannakoudakis2018developing} and LOCNESS \citep{granger1998computer} GED dataset\footnote{\url{https://www.cl.cam.ac.uk/research/nl/bea2019st/}} ($38,692$ sentences) released as part of the \textbf{BEA 2019} shared task \citep{bryant-etal-2019-bea}. It contains English essays written in response to varied topics and by English learners from different proficiency levels, as well as native English speakers. 
As the gold test set labels are not publicly available, we evaluate on the released development set and use $10\%$ of the training data for tuning\footnote{\url{https://github.com/bujol12/bert-seq-interpretability/blob/master/dev_indices_train_ABC.txt}}. 
For both GED datasets, we train the model to detect grammatically incorrect sentences and evaluate how well the methods can identify individual tokens that have been annotated as errors.

\begin{table*}[t]
\small
\centering
\setlength{\tabcolsep}{7.5pt}
\begin{tabular}{r|ccc|ccc|ccc}
\multicolumn{1}{c}{ }
 & \multicolumn{3}{c}{FCE} & \multicolumn{3}{c}{BEA 2019} & \multicolumn{3}{c}{CoNLL 2010} \\
 & Sent $F_1$  & $F_1$ & MAP & Sent $F_1$  & $F_1$ & MAP & Sent $F_1$ & $F_1$ & MAP\\
\hline
Random baseline & -  & 23.19 & 33.95 & -  & 16.73 & 27.01 & - & 1.63 & 14.15 \\
RoBERTa & 84.51 & - & - & 83.66 & - & - & 86.66 & - & - \\
\citet{rei-sogaard-2018-zero} & 84.75 & 28.73 & 48.56 & 81.27 & 18.53 & 31.69 & 84.16 & \textbf{72.42} & 87.82 \\ 
\hline
LIME & 84.51 & 24.60 & 37.90 & 83.66 & 2.09 & 31.41 & 86.66 & 57.14 & 78.44 \\
Attention heads & 84.51  & 24.34 & 48.04 & 83.66 & 19.69 & 40.55 & 86.66  & 25.64 & 79.82 \\
Soft attention & \textbf{85.62} & 32.16 & 48.90 & 83.41 & 22.92 & 35.79 & 86.25 & 8.45 & 20.04 \\
Weighted soft attention & \textbf{85.62}  & \textbf{33.31} & \textbf{53.91} & \textbf{83.68} & \textbf{24.35} & \textbf{41.07} & \textbf{87.20} & 67.28 & \textbf{91.18} \\ \hline
\end{tabular}
\caption{\label{res-fce-bea}
Results on FCE, BEA 2019 and CoNLL 2010. 
\textit{Sent} \textit{$F_1$} refers to F-measure on the sentence classification task; \textit{$F_1$} refers to token-level classification performance; \textit{MAP} is the token-level Mean Average Precision.
}
\end{table*}

\section{Experimental setup}

We use the pre-trained RoBERTa-base \citep{liu2019roberta} model, made available by HuggingFace \citep{wolf2019huggingface}, as our transformer architecture.
Following \citet{mosbach2020stability}, transformer models are fine-tuned for $20$ epochs, and the best performing checkpoint is then chosen based on sentence-level performance on the development set.
Each experiment is repeated with $5$ different random seeds and the averaged results are reported. 
The average duration of training on Nvidia GeForce RTX 2080Ti was $1$ hour. 
Significance testing is performed with a two-tailed paired t-test and $a = 0.05$. Hyperparameteres are tuned on the development set and presented in Appendices \ref{ap:hyper} and \ref{ap:threshold}. 

The {LIME} and {attention head} methods provide only a score without a natural decision boundary for classification. 
Therefore, we choose their thresholds based on the token-level $F_1$-score on the development set.
In contrast, the {soft attention} and {weighted soft attention} methods do not require such additional tuning that uses token-level labels.


\section{Results}
The results are presented in Table \ref{res-fce-bea}. 
Each model is trained as a sentence classifier and then evaluated as a token labeler. 
The challenge of the zero-shot sequence-labeling setting lies in the fact that the models are trained without utilizing any gold token-level signal; 
nevertheless, some methods perform considerably better than others. 
For reference, we also include a random baseline, which samples token-level scores from the standard uniform distribution; a RoBERTa model supervised as a sentence classifier only; and the model from \citet{rei-sogaard-2018-zero} based on BiLSTMs.

We report the $F_1$-measure on the token level along with Mean Average Precision {(MAP)} for returning positive tokens.  
The MAP metric views the task as a ranking problem and therefore removes the dependence on specific classification thresholds.
In addition, we report the $F_1$-measure on the main sentence-level task to ensure the proposed methods do not have adverse effects on sentence classification performance. Precision and recall values are included in Appendix \ref{ap:full_res}.

LIME has relatively low performance on FCE and BEA 2019, while it achieves somewhat higher results on CoNLL 2010. Comparing the MAP scores, the {attention head} method performs substantially better, especially considering that it is much more lightweight and requires no additional computation.
Nevertheless, both of these methods rely on using some annotated examples to tune their classification threshold,
which precludes their application in a truly zero-shot setting.

\begin{figure}[t]
  \centering
  \includegraphics[scale=0.15]{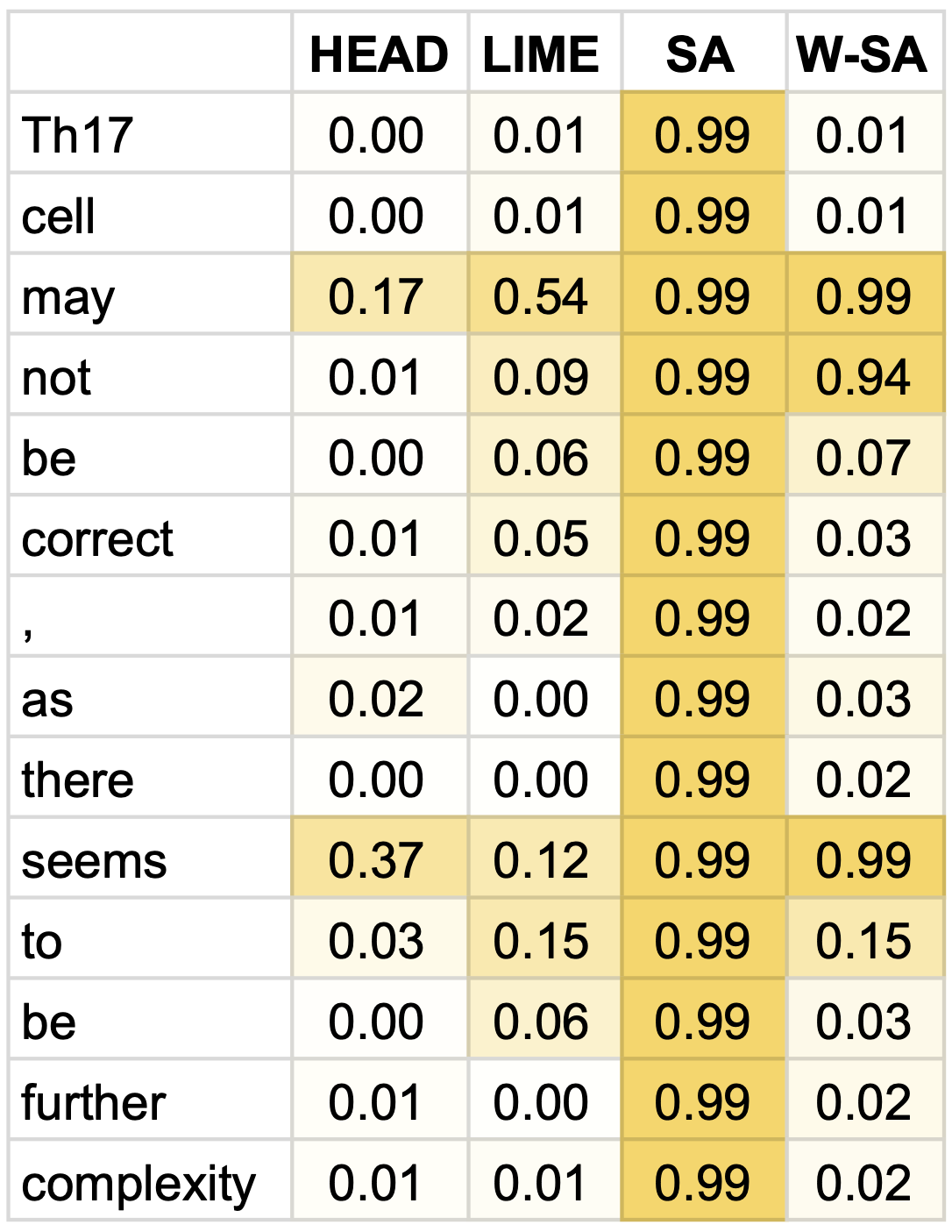}
  \caption{Example word-level importance scores $r_i$ (Eq. \ref{combine_tokens}) of different methods applied to an excerpt from the CoNLL10 dataset. \textit{HEAD} corresponds to {attention heads}; \textit{SA} to {soft attention}; and \textit{W-SA} to {weighted soft attention}. We can observe how \textit{W-SA} is the only method that correctly assigns substantially higher weights to the `may' and `seems' uncertainty cues.}
  \label{conll10_fig}
\end{figure}

Combining the soft attention mechanism with the transformer architecture provides some improvements over the previous methods, while also improving over \citet{rei-sogaard-2018-zero}. A notable exception is the CoNLL 2010 dataset where this method achieves only $8\%$ $F_1$ and $20\%$ MAP. Error analysis revealed that this is due 
to the transformer representations spreading attention scores evenly between a large number of tokens, as observed in Figure \ref{conll10_fig}.
Uncertainty cues in CoNLL 2010 can span across whole sentences (e.g., \textit{`Either ... or ...}'), with such examples encouraging the model to distribute information even further. 

The {weighted soft attention} modification addresses this issue and considerably improves performance across all metrics on all datasets.
Compared to the non-weighted version of the soft attention method, applying the extra weights leads to a significant improvement in terms of MAP, with a minimum of $5.01\%$ absolute gain on FCE.
The improvements are also statistically significant compared to the current state of the art \cite{rei-sogaard-2018-zero}: $5.35\%$ absolute improvement on FCE; $9.38\%$ on BEA 2019; and $3.36\%$ on CoNLL 2010. 
While the $F_1$ on CoNLL 2010 is slightly lower, the MAP score is higher, indicating that the model has difficulty finding an optimal decision boundary, but nevertheless provides a better ranking.
In future work, the weighted soft attention method for transformers could potentially be combined with token supervision in order to train robust multi-level models \cite{barrett2018sequence,rei2019jointly}.

\section{Conclusion}




We investigated methods for inferring token-level predictions from transformer models trained only on sentence-level annotations.
Experiments showed that previous approaches designed for LSTM architectures do not perform as well when applied to transformers. 
As transformer models already contain multiple layers of multi-head attention, the input representations get distributed between many tokens, making it more difficult to identify the importance of each individual token. LIME was not able to accurately identify target tokens, while the soft attention method primarily assigned equal attention scores across most words in a sentence.
Directly using the scores from the existing attention heads performed better than expected, but required some annotated data for tuning the decision threshold.
Modifying the soft attention module with an explicit sharpness constraint on the weights was found to encourage more distinct predictions, significantly improving token-level results. 

\section*{Acknowledgments}
We would like to thank James Thorne for his assistance in setting up the LIME experiments. Kamil Bujel was funded by the Undergraduate Research Opportunities Programme Bursary from the Department of Computing at Imperial College London.





\bibliographystyle{acl_natbib}
\bibliography{anthology,acl2021}

\clearpage
\appendix

\section{Example word-level predictions}
We present samples of  word-level predictions (word-level importance scores $r_i$, Eq. \ref{combine_tokens}) to illustrate differences between methods. In the figures that follow, \textit{HEAD} refers to {attention heads}, {SA} to {soft attention}, and \textit{W-SA} to {weighted soft attention}.
\label{ap:eg}

\begin{figure}[h!]
  \centering
  \includegraphics[scale=0.15]{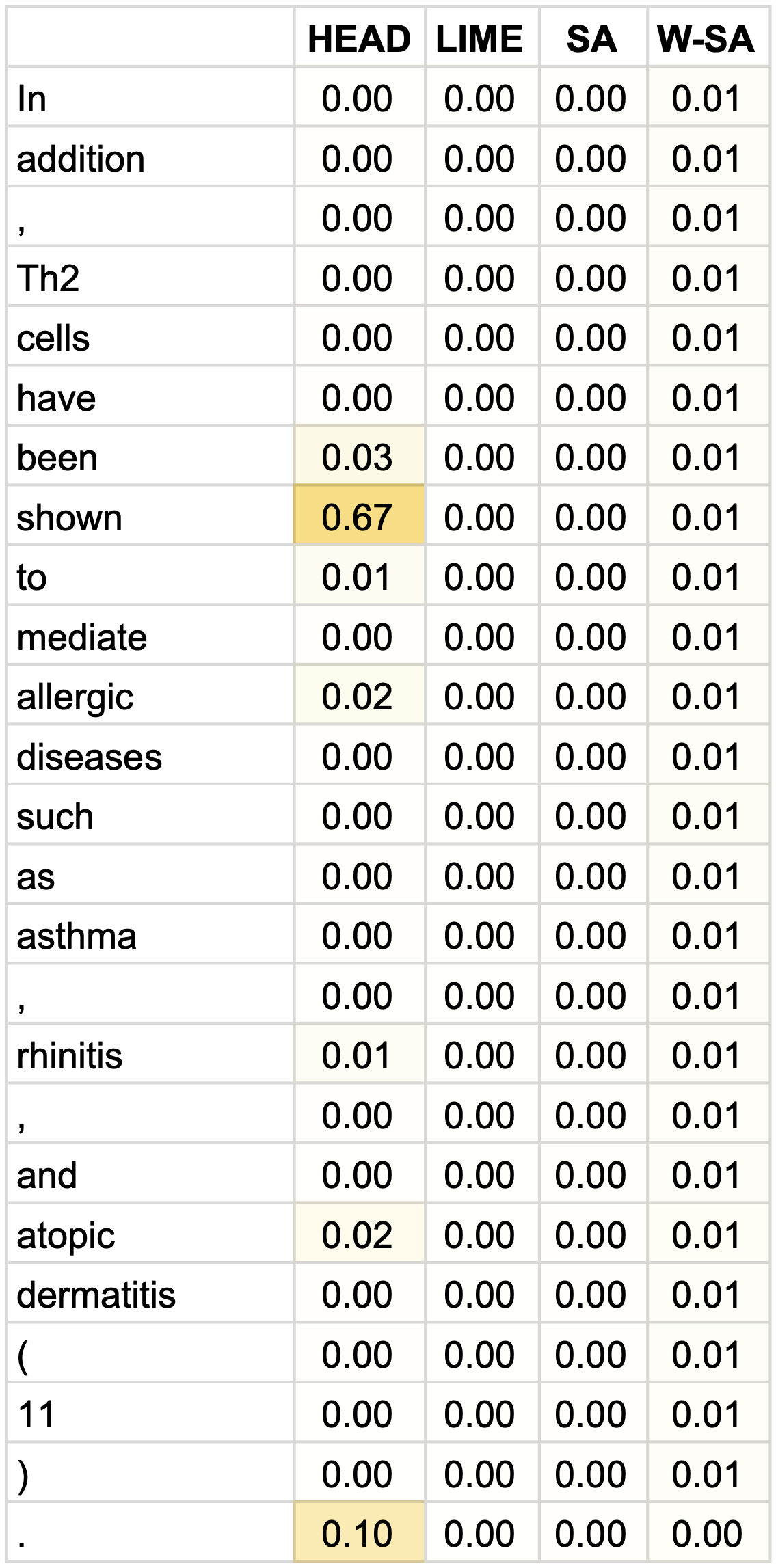}
  \caption{CoNLL 2010 negative sentence (without uncertainty cues). We can clearly see that most methods correctly put weights close to $0$ for all words, except from HEAD, which focuses on `shown' and `.'. We surmise this is due to the fact that, for HEAD, weights over the whole sentence have to sum up to $1$.} 
  \label{fig:sfig2}
\end{figure}

\begin{figure}[h!]
  \centering
  \includegraphics[scale=0.15]{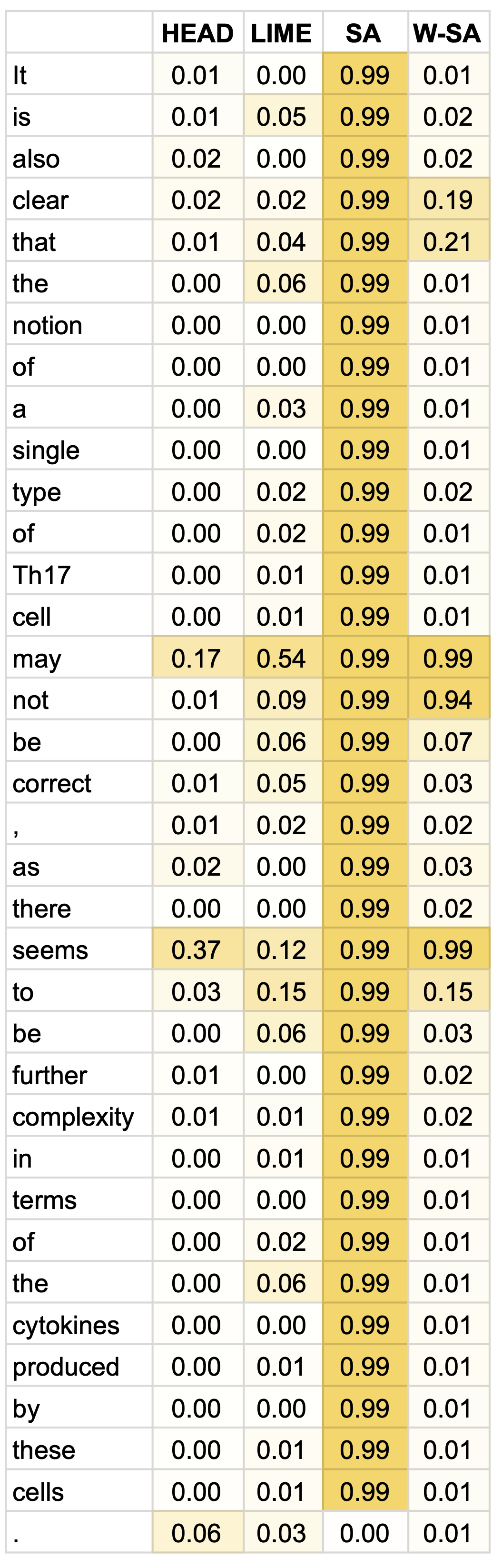}
  \caption{CoNLL 2010 positive sentence (with uncertainty cues). We can observe that HEAD  correctly identifies both of the uncertainty cues: `may' and `seems'; however the weight for `may' is quite low. Similarly, LIME identifies both tokens, but the weight for `seems' is particularly low (lower than for `to'). SA simply assigns high weights to all words. W-SA  focuses primarily on the two uncertainty cue words; however, it also incorrectly focuses on `not'.}
  \label{fig:sfig1}
\end{figure}

\begin{figure}[h!]
  \centering
  \includegraphics[scale=0.16]{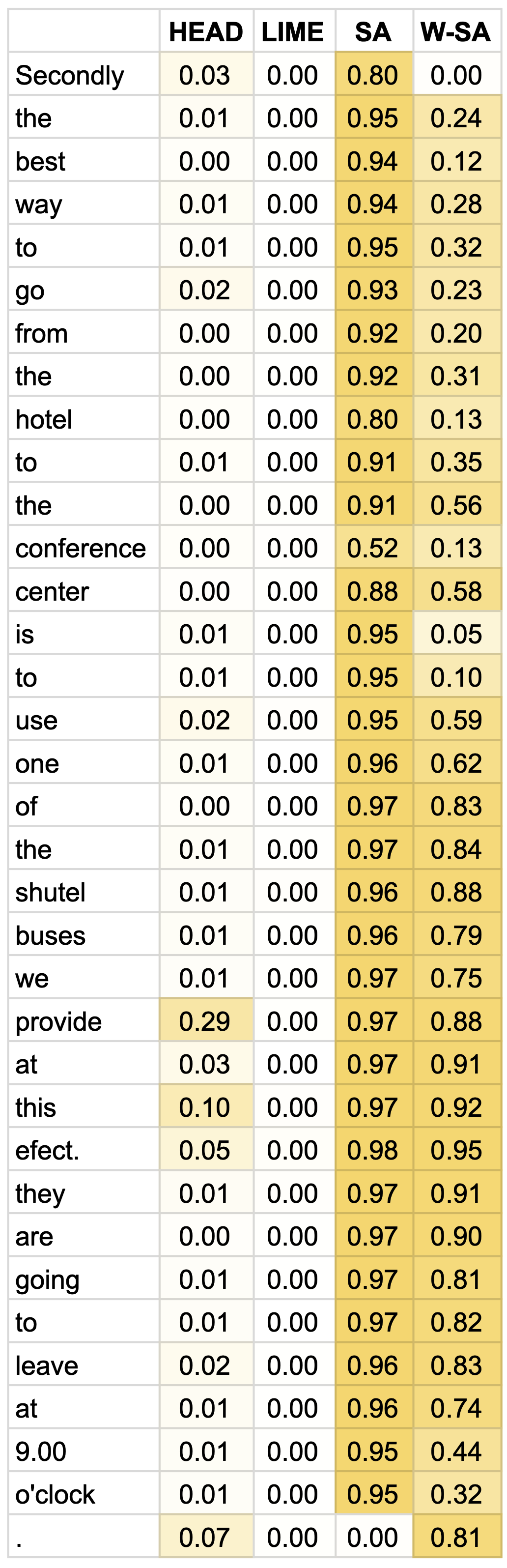}
  \caption{FCE positive sentence (contains grammatical errors). We can see that both LIME and HEAD struggle to assign informative and/or useful weights to the words. All SA weights are relatively high, with small variations in value. We can see that squaring (W-SA) leads to more well-defined weights over the whole sentence, with high weights mainly observed in the second part of the sentence, which is the one that contains incorrect words. However, on this dataset, even W-SA struggles to correctly identify which words precisely are incorrect.}
  \label{fig:sfig3}
\end{figure}

\section{Hyperparameters}
\label{ap:hyper}
\begin{table}[h]
\small
\centering
\begin{tabular}{l|c}
\textbf{Name} & \textbf{Value} \\
\hline
$\gamma$ & 0.1 \\
max seq length & 128 \\
per device train batch size & 16 \\
per device eval batch size & 64 \\
warmup ratio & 0.1 \\
learning rate & 2e-5 \\
weight decay & 0.1 \\
adam epsilon & 1e-7 \\
hidden layer dropout & 0.1 \\
soft attention layer size & 100 \\
soft attention hidden size & 300 \\
initializer & glorot \\
\end{tabular}
\caption{\label{Hyperparameters used during model training}
Model hyperparameters. 
}
\end{table}

\section{Word-level prediction thresholds}
\label{ap:threshold}
\begin{table}[h]
\small
\centering
\begin{tabular}{l|c|c}
\textbf{Dataset} & \textbf{Method} & \textbf{Threshold} \\
\hline
\textbf{CoNLL 2010} & LIME & 0.200 \\
 & Random baseline & 0.500 \\
 & Attention heads & 0.320 \\
 & \citet{rei-sogaard-2018-zero} & 0.500 \\
 & Soft attention & 0.500 \\
 & Weighted soft attention & 0.500 \\
\hline
\textbf{FCE} & LIME & 0.001 \\
 & Random baseline & 0.500 \\
 & Attention heads & 0.080 \\
 & \citet{rei-sogaard-2018-zero} & 0.500 \\
 & Soft attention & 0.500 \\
 & Weighted soft attention & 0.500 \\
\hline
\textbf{BEA 2019} 
 & LIME & 0.010 \\
  & Random baseline & 0.500 \\
 & Attention heads & 0.080 \\
 
 & \citet{rei-sogaard-2018-zero} & 0.500 \\
 & Soft attention & 0.500 \\
 & Weighted soft attention & 0.500 \\
\end{tabular}
\caption{Word-level thresholds above which a word is classified as positive.
}
\end{table}

\section{Validation set results}
\begin{table}[h!]
\small
\centering
\begin{tabular}{l|c|c}
\textbf{Dataset} & \textbf{Method} & \textbf{Sent $F_1$} \\
\hline
\textbf{CoNLL 2010} & LIME & 91.77 \\
 & RoBERTa & 91.77 \\
 & Attention heads & 91.77 \\
 & Soft attention & 92.12 \\
 & Weighted soft attention & 91.82 \\
\hline
\textbf{FCE} & LIME & 84.49 \\
 & RoBERTa & 84.49 \\
 & Attention heads & 84.49 \\
 & Soft attention & 84.82 \\
 & Weighted soft attention & 85.56 \\
\hline
\textbf{BEA 2019} & LIME & 83.65 \\
 & RoBERTa & 83.65 \\
 & Attention heads & 83.65 \\
 & Soft attention & 83.47 \\
 & Weighted soft attention & 83.64 \\
\end{tabular}
\caption{Mean sentence-level $F_1$ score on the development set, averaged over $5$ runs.
}
\end{table}

\clearpage

\section{Full test set results}
\label{ap:full_res}
\begin{table}[!h]
\small
\centering
\setlength{\tabcolsep}{2pt}
\begin{tabular}{r|ccc}
 & \multicolumn{3}{c}{FCE}  \\
 & Sent $F_1$ & Sent P & Sent R \\
\hline
Random baseline & - & - & -  \\
RoBERTa & 84.51 &  84.25 & \textbf{84.93} \\
\citet{rei-sogaard-2018-zero} & 84.75 & - & - \\ \hline
LIME & 84.51 &  84.25 & \textbf{84.93} \\
Attention heads & 84.51 & 84.25 & \textbf{84.93}   \\
Soft attention & \textbf{85.62} & \textbf{86.92} & 84.42\\
Weighted soft attention & \textbf{85.62} & 86.88 & 84.45 \\
\end{tabular}
\caption{
Sentence-level results: \textit{P}, \textit{R} and \textit{$F_1$} refer to Precision, Recall and F-measure respectively on the positive class.
}
\end{table}

\begin{table}[!h]
\small
\centering
\setlength{\tabcolsep}{2pt}
\begin{tabular}{r|ccc}
 & \multicolumn{3}{c}{BEA 2019}  \\
 & Sent $F_1$ & Sent P & Sent R \\
\hline
Random baseline & - & - & -  \\
RoBERTa & 83.66 & \textbf{82.29} & 85.15 \\
\citet{rei-sogaard-2018-zero} & 81.27 & - & - \\ 
\hline
LIME & 83.66 & \textbf{82.29} & 85.15 \\
Attention heads & 83.66 & \textbf{82.29} & 85.15 \\
Soft attention &  83.41 & 81.47 & 85.54 \\

Weighted soft attention & \textbf{83.68} & 79.95 &  \textbf{87.91} \\
\end{tabular}
\caption{
Sentence-level results: \textit{P}, \textit{R} and \textit{$F_1$} refer to Precision, Recall and F-measure respectively on the positive class.
}
\end{table}

\begin{table}[!h]
\small
\centering
\setlength{\tabcolsep}{2pt}
\begin{tabular}{r|ccc}

 & \multicolumn{3}{c}{CoNLL 2010}  \\
 & Sent $F_1$ & Sent P & Sent R \\
\hline
Random baseline &  - & - & - \\
RoBERTa & 86.66 & 84.90 & \textbf{88.63} \\
\citet{rei-sogaard-2018-zero} & 84.16 & - & -  \\ 
\hline
LIME & 86.66 & 84.90 & \textbf{88.63} \\
Attention heads  & 86.66 & 84.90 & \textbf{88.63} \\

Soft attention & 86.25 & 85.75	& 86.89  \\

Weighted soft attention & \textbf{87.20} & \textbf{89.17} & 85.37 \\
\end{tabular}
\caption{
    Sentence-level results: \textit{P}, \textit{R} and \textit{$F_1$} refer to Precision, Recall and F-measure respectively on the positive class.
}
\end{table}

\newpage

\begin{table}[!h]
\small
\centering
\setlength{\tabcolsep}{2pt}
\begin{tabular}{r|cccc}
 & \multicolumn{4}{c}{FCE}  \\ & P & R & $F_1$ & MAP \\
\hline
Random baseline  & 15.11 & 49.81 & 23.19 & 33.95  \\
RoBERTa & - & - & - & - \\
\citet{rei-sogaard-2018-zero}  & \textbf{29.16} & 29.04 & 28.73 & 48.56 \\ \hline
LIME & 19.06 & 34.70 & 24.60 & 37.90 \\
Attention heads & 26.67 & 22.38 & 24.34 & 48.04   \\
Soft attention & 19.84 & \textbf{85.38} & 32.16 & 48.90 \\
Weighted soft attention & 20.76 & 85.36 & \textbf{33.31} & \textbf{53.91} \\
\end{tabular}
\caption{
Token-level results: \textit{P}, \textit{R} and \textit{$F_1$} refer to Precision, Recall and F-measure respectively on the positive class. \textit{MAP} is the Mean Average Precision at the token-level.
}
\end{table}

\begin{table}[!h]
\small
\centering
\setlength{\tabcolsep}{2pt}
\begin{tabular}{r|cccc}
 & \multicolumn{4}{c}{BEA 2019}  \\
& P & R & $F_1$ & MAP \\
\hline
Random baseline & 10.05 & 50.00 & 16.73 & 27.01  \\
RoBERTa & - & - & - & - \\
\citet{rei-sogaard-2018-zero} & 10.93 & 61.63 & 18.53 & 31.69 \\ 
\hline
LIME  & 13.49 & 1.13 & 2.09 & 31.41 \\
Attention heads & \textbf{18.48} & 21.07 & 19.69 & 40.55 \\
Soft attention & 13.20 & \textbf{87.19} & 22.92 & 35.79 \\

Weighted soft attention & 14.20 & 85.49 & \textbf{24.35} & \textbf{41.07} \\
\end{tabular}
\caption{
Token-level results: \textit{P}, \textit{R} and \textit{$F_1$} refer to Precision, Recall and F-measure respectively on the positive class. \textit{MAP} is the Mean Average Precision at the token-level.
}
\end{table}

\begin{table}[!h]
\small
\centering
\setlength{\tabcolsep}{2pt}
\begin{tabular}{r|cccc}

 & \multicolumn{4}{c}{CoNLL 2010}  \\
 & P & R & $F_1$ & MAP \\
\hline
Random baseline & 0.83 & 49.70 & 1.63 & 14.15 \\
RoBERTa & - & - & - & - \\
\citet{rei-sogaard-2018-zero} & \textbf{78.99} & 67.06 & \textbf{72.42} & 87.82 \\ 
\hline
LIME & 63.25 & 52.11 & 57.14 & 78.44 \\
Attention heads & 22.33 & 30.11 & 25.64 & 79.82 \\

Soft attention & 4.48 & \textbf{86.14} & 8.45 & 20.04 \\

Weighted soft attention & 58.80 & 78.89 & 67.28 & \textbf{91.18} \\
\end{tabular}
\caption{
    Token-level results: \textit{P}, \textit{R} and \textit{$F_1$} refer to Precision, Recall and F-measure respectively on the positive class. \textit{MAP} is the Mean Average Precision at the token-level.
}
\end{table}

\clearpage
\section{Weighted soft attention architecture}

\begin{minipage}{\textwidth}
    \label{fig:sa-diag}
    \includegraphics[width=0.90\textwidth,height=10cm]{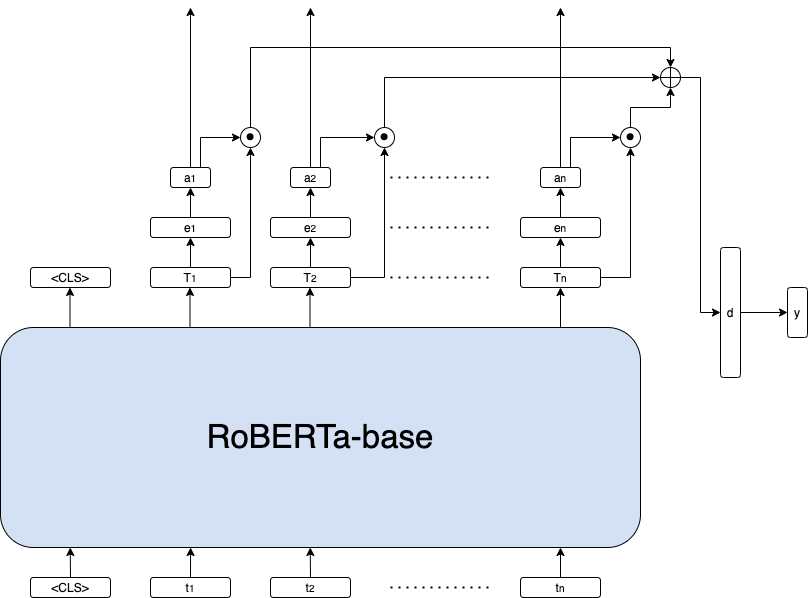}
    \captionof{figure}{Architecture of our proposed weighted soft attention model. $[t_1, t_2, ..., t_n]$ represent the tokenized input sentence, while $[T_1, T_2, ..., T_n]$ are the resulting contextual embeddings. $[e_1, e_2, ..., e_n]$ are attention vectors, and $[a_1, a_2, ..., a_n]$ are normalized attention weights. $d$ represents the output vector and $y$ the final output logits.}
\end{minipage}

\end{document}